\documentclass[letterpaper, 10 pt, journal, twoside]{IEEEtran}

\usepackage{cite}
\usepackage{booktabs}
\usepackage[pdftex]{graphicx}
\usepackage[export]{adjustbox}
\usepackage{svg}
\usepackage{tikz}
\usepackage{xcolor}
\usepackage{subcaption}
\usepackage{multirow}
\usepackage{titlesec}
\IEEEaftertitletext{\vspace{-20pt}}
\usepackage{gensymb}
\usepackage{amsmath}
\usepackage{amssymb}
\usepackage{hyperref}
\newcommand{\State}{\mathcal{S}}

\newcommand{\StatePT}{\mathcal{S}_{\text{PT}}}
\newcommand{\StateHA}{\mathcal{S}_{\text{HA}}}

\begin{document}

\title{DR-MPC: Deep Residual Model Predictive Control for Real-world Social Navigation}

\author{James~R.~Han$^{1}$, Hugues~Thomas$^{2}$, Jian~Zhang$^{2}$, Nicholas Rhinehart$^{1}$, Timothy~D.~Barfoot$^{1}$
\thanks{Manuscript received: October, 11, 2024; Revised January, 2, 2025; Accepted February, 4, 2025.}
\thanks{This paper was recommended for publication by Editor Angelika Peer upon evaluation of the Associate Editor and Reviewers' comments.} 
\thanks{\textsuperscript{1} James R. Han, Nicholas Rhinehart, and Timothy D. Barfoot are with the University of Toronto Institute for Aerospace Studies, Canada {\tt\footnotesize jamesr.han@mail.utoronto.ca, nick.rhinehart@utoronto.ca, tim.barfoot@utoronto.ca}}
\thanks{\textsuperscript{2} Hugues Thomas and Jian Zhang are with Apple, USA {\tt\footnotesize hugues.thomas@robotics.utias.utoronto.ca, air23zj@gmail.com}}
\thanks{Digital Object Identifier (DOI): see top of this page.}
}

\markboth{IEEE Robotics and Automation Letters. Preprint Version. Accepted February, 2025}
{Han \MakeLowercase{\textit{et al.}}: DR-MPC}

\maketitle

\begin{abstract}
How can a robot safely navigate around people with complex motion patterns? Deep Reinforcement Learning (DRL) in simulation holds some promise, but much prior work relies on simulators that fail to capture the nuances of real human motion. Thus, we propose Deep Residual Model Predictive Control (DR-MPC) to enable robots to quickly and safely perform DRL from real-world crowd navigation data. By blending MPC with model-free DRL, DR-MPC overcomes the DRL challenges of large data requirements and unsafe initial behavior. DR-MPC is initialized with MPC-based path tracking, and gradually learns to interact more effectively with humans. To further accelerate learning, a safety component estimates out-of-distribution states to guide the robot away from likely collisions. In simulation, we show that DR-MPC substantially outperforms prior work, including traditional DRL and residual DRL models. Hardware experiments show our approach successfully enables a robot to navigate a variety of crowded situations with few errors using less than 4 hours of training data (video: \url{https://youtu.be/GUZlGBk60uY}, code: \url{https://github.com/James-R-Han/DR-MPC}).
\end{abstract}


\IEEEpeerreviewmaketitle

\vspace{-5mm}
\section{Introduction}

\IEEEPARstart{A}{chieving} reliable robotic navigation remains one of the main challenges in integrating mobile robots into society. Applications such as food service and equipment transport could see major cost and time savings but require robots to navigate human environments with static obstacles.

Social robot navigation research focuses on enabling robots to move safely and efficiently around humans. Deep Reinforcement Learning (DRL) has emerged as a promising alternative to traditional learning-free approaches including reactive, decoupled, and coupled planning. Reactive methods model humans as static objects \cite{DWA}, decoupled planning approaches forecast future human trajectories to generate a cost map for navigation \cite{HuguesForeseeableFuture}, and coupled planning approaches model human motion and solve a joint optimization problem \cite{SICNav}. These methods have notable shortcomings. Reactive approaches are shortsighted, resulting in intrusive behaviour \cite{HuguesForeseeableFuture}. Decoupled approaches neglect the robot's impact on humans, which can cause the robot to `freeze' when the robot's plan conflicts with forecasted human trajectories \cite{EvolutionSociallyAwareRobot2023}. Lastly, coupled methods falter when the human model is inaccurate \cite{SICNav}, and accurately modeling human motion is challenging. 

DRL offers a compelling value proposition: learn efficient and safe robot behaviour without a human motion model. Current research centers around simulation due to the dangers posed by randomly initialized DRL agents and the extensive training data requirement \cite{RealWorldRLDifficulty}. However, these simulators use human models that are mismatched with the true human motion, making sim-to-real agents unfit for deployment. For instance, the popular CrowdNav simulator assumes a cooperative and deterministic human policy, \cite{CrowdNavCrowdSim} Optimal Reciprocal Collision Avoidance (ORCA) \cite{ORCA}, which leads to aggressive DRL agents \cite{DSRNN}. Also, many simulators neglect the presence of static obstacles, further exacerbating the sim-to-real gap.

\begin{figure}
  \centering
  \includegraphics[width=0.47\textwidth]{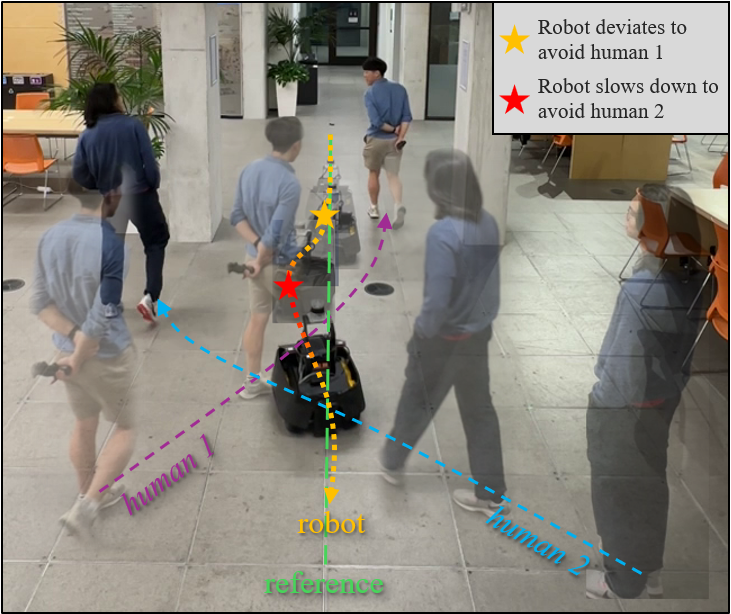} 
  \caption{DR-MPC navigating in the real world. In this illustration, the robot deviates from its path to allow human 1 to pass and then slows down (red means slower speed) to let human 2 to pass before returning to its path.}
  \label{fig:RealWorldMoneyShot}
\end{figure}

\begin{figure*}[htbp]
    \centering
    \includegraphics[width=1.0\textwidth]{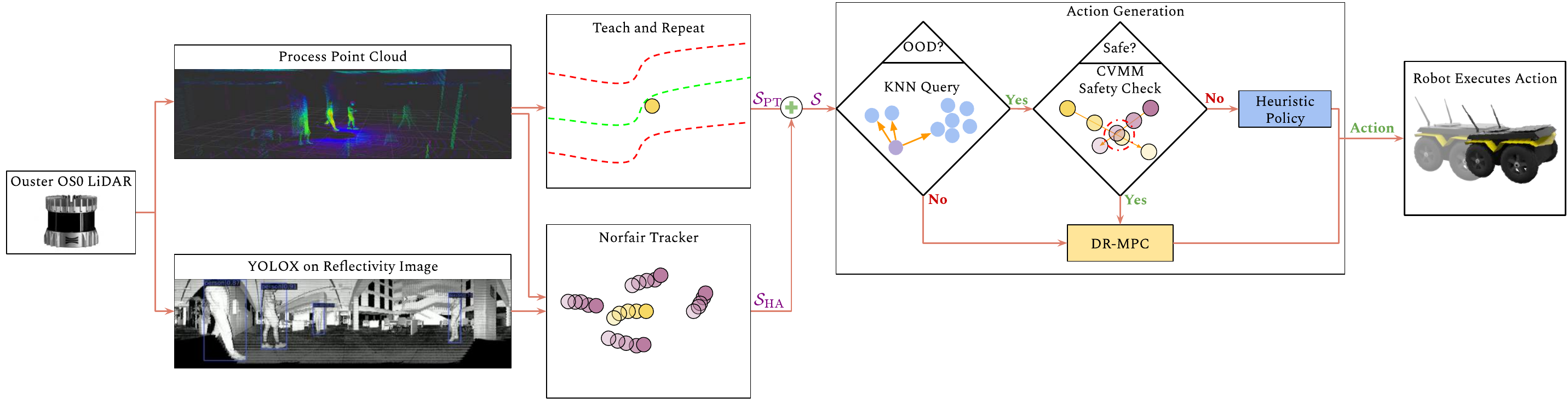}
    \caption{Full real-world pipeline. The Ouster OS0-128 LiDAR generates a detailed reflectivity image and a point cloud. The reflectivity image allows us to perform human tracking and the point cloud enables localization, path tracking, and depth recovery. With the state constructed from a single sensor, we use our OOD module and CVMM safety check to determine whether or not to execute the DR-MPC or the heuristic safety policy.}
    \label{fig:Full Pipeline}
\end{figure*}

In this paper, we train a DRL agent directly in the real world. To handle environments with static obstacles, we modify the typical DRL social navigation Markov Decision Process (MDP) into human avoidance with path tracking within virtual corridors. We introduce a novel approach, Deep Residual Model Predictive Control (DR-MPC), that integrates Model Predictive Control (MPC) path tracking to significantly accelerate the learning process. Lastly, we design a pipeline with out-of-distribution (OOD) state detection and a heuristic policy to guide the DRL agent into higher-reward regions. Our approach enabled real-world deployment of a DRL agent without any simulation with less than 4 hours of data.

\section{Related Works}

\textbf{Social Navigation Simulators}. Several 2D and 3D simulators continue to be developed for social navigation. These simulators offer a way to safely test and experiment with different approaches. Simulators also offer the possibility, in theory, of developing `sim-to-real' approaches, which can substantially reduce the amount of real-world learning (and mistakes) that robots make. A popular simulator for social navigation is the 2D CrowdNav simulator \cite{CrowdNavCrowdSim}, designed for waypoint navigation in open spaces with the human policy defined by ORCA \cite{ORCA}. While CrowdNav has facilitated substantial DRL model development \cite{RGL, DSRNN, IntentionAwareGraph, NaviSTAR}, the cooperative nature of the ORCA policy creates a large sim-to-real gap. When the robot is visible to humans, the learned DRL policy is dangerously aggressive: the robot pushes humans out of its way to reach its goal \cite{DSRNN}. Consequently, the {\em invisible testbed}---when the robot is invisible to the humans---was adopted as the benchmark standard. Although the invisible testbed reduces the DRL agent's aggression, the problem becomes equivalent to decoupled planning where the DRL agent does not learn how its action will influence humans \cite{DSRNN}.  

Beyond CrowdNav, other simulators use human motion models such as variations of the Social Forces Model (SFM) or behaviour graphs \cite{Arena3, HuNavSim}. Unfortunately, even the latest simulators struggle with human realism and often exhibit unsmooth human motions and enter deadlock scenarios. These deficiencies create a significant sim-to-real gap, where models trained in simulation often perform differently--—and usually worse--—when applied in the real world \cite{SimToRealChallenge}. While simulators are essential tools for advancing DRL model architectures, the most accurate data for DRL social navigation is real-world data. By reducing the amount of training data required, we enable the direct training of DR-MPC in the real world, avoiding unnecessary inductive biases introduced by simulators.

\begin{figure*}[htbp]
  \centering
  \includegraphics[width=0.95\textwidth]{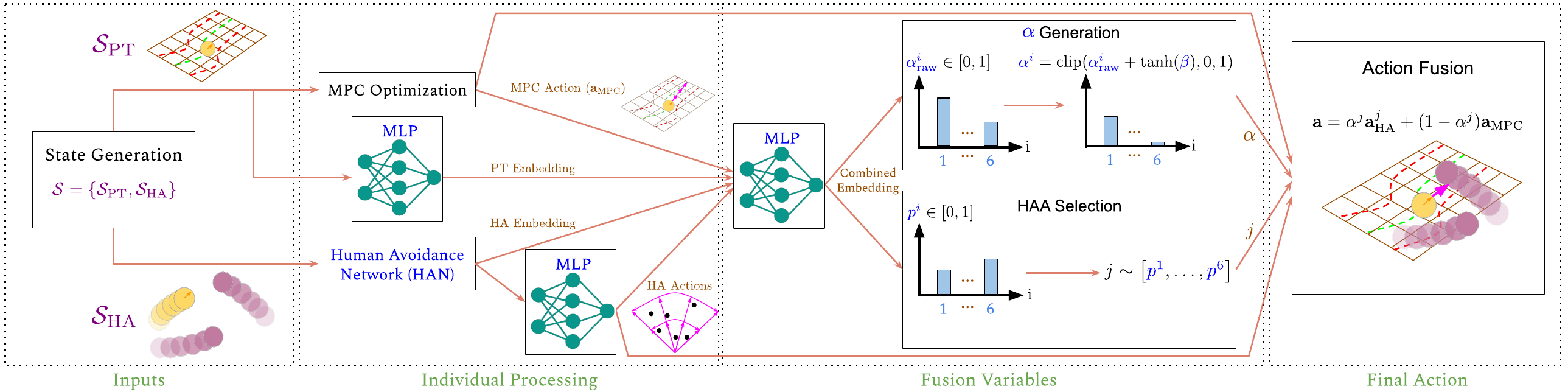}
  \caption{DR-MPC architecture. The dark blue text elements involve learning. From $\StatePT$, we generate the MPC path tracking action and a latent embedding of the path information using an MLP. From $\StateHA$, we use a SOTA human avoidance network to generate six actions for human avoidance. The model then fuses all the information to generate $\boldsymbol{\alpha}$ and $\mathbf{p}$, which generates the final action to maximize the human avoidance and path tracking rewards.}
  \label{fig:DR-MPC}
\end{figure*}

\textbf{RL Social Navigation Models}. Crowd navigation, a major sub-field of DRL social navigation, focuses on enabling a robot to navigate among humans in open areas. Significant progress has been made in incorporating machine learning advancements to enhance model reasoning about crowds in spatial and temporal dimensions.

Early models analyzed the scene by considering humans individually \cite{CrowdNavDecentralized2016, chenSociallyAwareMotion2017}. Then, approaches incorporated attention mechanisms to reason about the crowd as a whole, using learned attention scores to generate a crowd embedding for decision making \cite{CrowdNavCrowdSim, DSARL, CrowdNavGCNGaze}. Subsequent advances analyzed both human-robot and human-human interactions through graph neural networks (GNNs) \cite{RGL, CrowdNavGCNGaze, IntentionAwareGraph, NAX}. Most recently, models include temporal reasoning using spatio-temporal graphs, multihead attention mechanisms, and transformers \cite{DSRNN, NaviSTAR}, enabling reasoning on the trajectory level.

We do not focus on developing a novel architecture for waypoint-based crowd navigation; we incorporate state-of-the-art (SOTA) architectures for processing humans into DR-MPC.

\textbf{Residual DRL}. Residual DRL integrates a user-supplied and a learned policy \cite{ResidualRLForRobotics}. DRL learns corrective actions on top of the base controller. Learning speedup occurs because on initialization the model has an expected action equal to the base controller, which, assuming a suitable base controller, results in performance better than a random policy \cite{ResidualRLForRobotics}. 

To the best of our knowledge, residual DRL has not been applied to social navigation, but some works have attempted to incorporate classical control. Kästner et al. \cite{DRLSwitchController} learn a DRL policy that switches between different controllers, each optimized for an individual task. Another work, Semnani et al. \cite{DRL-FMP} switch from a DRL policy to a force-based model when close to humans. While these approaches reduce DRL's learning burden, the overall performance is limited by the base controllers. When the policy switches to a base controller, DRL can not optimize the individual controllers.

In contrast, DR-MPC fully exploits the capabilities of MPC path tracking, which is an optimal behaviour when no humans are present. Like residual DRL, DR-MPC can replicate the MPC action or generate an action far from it. Unlike residual DRL, DR-MPC's initial behaviour almost exactly follows MPC path tracking, the best performance without any prior human information. Additionally, DR-MPC learns to dynamically integrate or disregard the MPC action, leading to significantly faster training compared to residual DRL.

\textbf{Path Tracking}. DRL is beneficial for path tracking when modeling difficulties arise, but in indoor environments with minimal disturbances and a robot that can be accurately modeled with unicycle kinematics, we can leverage MPC for path tracking using the formulation presented by Sehn et al. \cite{JordyMPC}. As a result, DRL does not need to explicitly learn path tracking, thereby reducing the overall learning complexity.

\section{Decision process formulation} \label{sec:The MDP}
We modify the social navigation MDP in \cite{CrowdNavCrowdSim} to combine human avoidance and path tracking within virtual corridors, where these corridors ensure safety from static obstacles. To simplify the problem, we assume a constant corridor width.

\textbf{State Space}. We construct our state $\State = \{ \StatePT, \StateHA \}$, where $\StatePT$ is the state information for path tracking and $\StateHA$ is the state information for human avoidance.

As in \cite{DSRNN}, we exclude human velocities due to the difficulty of estimating these quantities in the real world and assume a constant human radius. If at time $t$ there are $n_t$ visible humans: $\StateHA = \left\{
    \mathbf{v}^{t-H:t-1}, \mathbf{r}^{t-H:t}, \mathbf{q}^{t-H_1:t}_1, \dots, \mathbf{q}^{t-H_{n_t}:t}_{n_t}
\right\}$, where $\mathbf{v}^{t-H:t-1}$ is the robot's past velocities from time $t-H$ to $t-1$, $\mathbf{r}^{t-H:t}$ is the robot's past positions in the current robot frame from time $t-H$ to $t$, and $\mathbf{q}^{t-H_{i}:t}_{i}$ is the $i^{\text{th}}$ visible human's past positions in the current robot frame from time $t-H_{i}$ to $t$, capped at length $H$.

We select a local path representation as in \cite{DATT}. $\StatePT$ includes the path node closest to the robot, along with the preceding $L$ nodes and the following $F$ nodes. All nodes are transformed into the robot's current reference frame.

\textbf{Actions}. The action space is a linear and angular velocity: $\mathbf{a} = \begin{bmatrix}
    v \hspace{0.2cm} \omega
\end{bmatrix}$.

\textbf{Rewards}. Our reward function considers path advancement, path deviation, goal reaching, corridor collisions, small speeds, human collisions, and human disturbance:
\begin{equation}
    r = r_\text{pa} + r_\text{dev} + r_\text{goal}^* + r_\text{cor-col}^* + r_\text{act}^* + r_\text{hum-col}^* + r_\text{dist},
\end{equation}

where rewards marked with an asterisk are terminal rewards. Path advancement: 
$r_\text{pa} = 5 \Delta s$ where $\Delta s$ is the arclength progress along the path.
Deviation: $r_\text{dev} = 0.5 d_\text{xy} + 0.03 |d_\theta|$ where $d_\text{xy}$ is the Euclidean distance and $d_\theta$ is the angular offset from the closest point on the path.
Goal: $r_\text{goal}^* = -5$ if the heading difference exceeds a threshold and 0 otherwise.
Corridor collision: $r_\text{cor-col}^* = -10$
for colliding with the corridor. 
Minimal actuation: $r_\text{act}^* = -20$ if the sum of the robot's past $H$ speeds falls below a threshold.

For human avoidance, we align our rewards with the two most important principles of human-robot interaction (HRI): safety and comfort \cite{SocialRobotEvaluation}. For safety, human collision: $r_\text{hum-col}^* = -15$.
For comfort, disturbance penalty \cite{NAX}: 
$r_\text{dist} = - \sum_{i=1}^{n_t} \left( 5.6 |\Delta v_\text{h}^i| + 3.5 |\Delta \theta_\text{h}^i| \right)$,
which penalizes the robot for causing changes in a human's velocity ($\Delta v_\text{h}^i$) and heading direction ($\Delta \theta_\text{h}^i$). These changes are computed relative to the human's velocity and heading at the previous time step.

Finally, we add two safety layers: safety-human and safety-corridor raises, which are conservative versions of the human-collision and corridor-collision penalties. While a safety violation does not guarantee a collision, it is likely. So, we slightly reduce the theoretical performance limit for safety.

\section{DR-MPC Policy Architecture} \label{sec:DR-MPC Model Architecture}

DR-MPC (Figure \ref{fig:DR-MPC}) consists of two main components: individual processing of $\StatePT$ and $\StateHA$, and the information fusion to generate a single action. We generate a latent embedding of the path using a Multi-layer Perceptron (MLP), and the MPC optimization from \cite{JordyMPC} generates $\mathbf{a}_\text{MPC}$. 

To process $\StateHA$, we modify \cite{IntentionAwareGraph} to handle varying-length human trajectories for off-policy learning. We refer to this adapted architecture as the Human Avoidance Network (HAN); further details are provided in the appendix. The output of HAN is a `crowd embedding' that is used to generate six candidate human avoidance actions ($\mathbf{a}_{\text{HA}}$), where each $\mathbf{a}^\text{i}_{\text{HA}}$ is positioned in a different cell of the action space. The human-avoidance action space is partitioned into six cells defined by two linear velocity bins $[v_{\text{min}}, v_{\text{middle}}]$, $[v_{\text{middle}}, v_{\text{max}}]$, and three angular velocity bins $[w_{\text{min}}, w_{\text{lower}}]$, $[\omega_{\text{lower}}, \omega_{\text{upper}}]$, $[\omega_{\text{upper}}, \omega_{\text{max}}]$. We include multiple actions because often several viable actions exist for human avoidance. We found empirically that this design reduces learning time compared to having the model learn the diversity. We also found that using six actions strikes a good balance between sector granularity and the data required to adequately sample and explore each sector. The output of the MLP following HAN is the mean action within each cell. Using a predetermined variance that decays over time, we sample a Gaussian to get $\mathbf{a}^\text{i}_{\text{HA}}$; this standard formulation comes from \cite{TD3}.

The key innovation of our model lies in how we combine these individual components. The path tracking embedding, the crowd embedding, $\mathbf{a}_\text{MPC}$, and $\mathbf{a}_{\text{HA}}$ are combined to output $\boldsymbol{\alpha}_\text{raw} = \begin{bmatrix}
    \alpha^1_\text{raw} \dots \alpha^6_\text{raw}
\end{bmatrix}$, where each $\alpha^i_\text{raw} \in [0,1]$, and a categorical distribution $\mathbf{p} = \begin{bmatrix}
    p^1 \dots p^6
\end{bmatrix}$. Note, as in Soft Actor-Critic (SAC), the model learns the mean and log-standard deviation to generate $\alpha^i_\text{raw}$, which is then passed through a $\tanh$ function to squash it, followed by scaling and shifting \cite{SACAlgorithms}. We then compute $\alpha^i = \text{clip}(\alpha^i_\text{raw} + \tanh(\beta),0,1)$, where $\beta$ is a learned parameter. Each $\mathbf{a}^\text{i}_{\text{HA}}$ corresponds to $\alpha^i$ and $p^i$ of the same index. After sampling the index $j$ from $\mathbf{p}$, the final action is constructed as a weighted sum: 
$\mathbf{a} = \alpha^j \mathbf{a}^\text{j}_\text{HA} + (1-\alpha^j)\mathbf{a}_\text{MPC}$.

Unlike residual DRL, DR-MPC begins with near-MPC behaviour by biasing actions toward MPC, initializing $\beta = -0.8$ to suppress $\boldsymbol{\alpha}$. Note that initializing $\beta$ too low yields little qualitative difference and requires a lot of model updates to raise $\beta$ towards 0. We observe the model naturally learns to adjust $\beta$ towards 0 to be able to take non-MPC actions.

We train our model with Discrete Soft Actor-Critic (DSAC) with double average clipped Q-learning with Q-clip to properly backpropagate through $\mathbf{p}$ \cite{RevisitingDSAC}. We augment DSAC with the entropy formulation from SAC to optimize the log-standard deviation that generates $\alpha^i_\text{raw}$ \cite{SACAlgorithms}. Figure \ref{fig:DR-MPC} depicts the actor. The critic shares the same architecture up to and including the first MLP in the fusion stage; however, this MLP outputs the Q-value, and the action is included as input.

\section{The Pipeline}\label{sec:The Pipeline}
To perform better than random exploration, we guide the selection of $\mathbf{a}^i_{HA}$. Early intervention in uncertain or high-risk states has proven effective in interactive imitation learning \cite{hgdagger}. We perform OOD state detection: if the state is in-distribution (ID), we execute DR-MPC's action. Otherwise, we either validate the model's action as safe using a heuristic safety check or override it using a heuristic policy to steer the robot toward collision-free areas.

Once a state is ID, the DRL agent will continue to explore and may still encounter collisions. The agent will quickly learn to focus on known good actions and avoid poor reward regions, thereby accelerating learning speed. Note that the OOD state detection, heuristic safety check, and heuristic policy are modular components and can be swapped out or fine-tuned independently of everything else in the pipeline.

\subsubsection{OOD State Detection}
We use the SOTA OOD algorithm: $K$-nearest neighbors (KNN) in the latent space of our model \cite{OODKNN}. We extract the model's latent embedding for each state in the replay buffer. With the Faiss package \cite{faiss}, we can efficiently query the $K$'th closest vector. A state is considered OOD if its distance to the $K$'th nearest vector exceeds the threshold: the average distance between ($\State$, $\State'$) pairs in the replay buffer. Intuitively, if a query state is ID, it will roughly have $K$ other states nearby. This method offers two key advantages over approaches like Random Network Distillation (RND) \cite{RNDOODSN}. First, it requires minimal additional compute as it leverages the existing model's latent space without needing to train a new model. Second, by explicitly utilizing all the data, it avoids capacity issues where methods such as RND can `forget' due to the model's limited capacity.

\subsubsection{Heuristic Safety Check and Policy}
\begin{figure}
    \centering
    \includegraphics[width=0.4\textwidth]{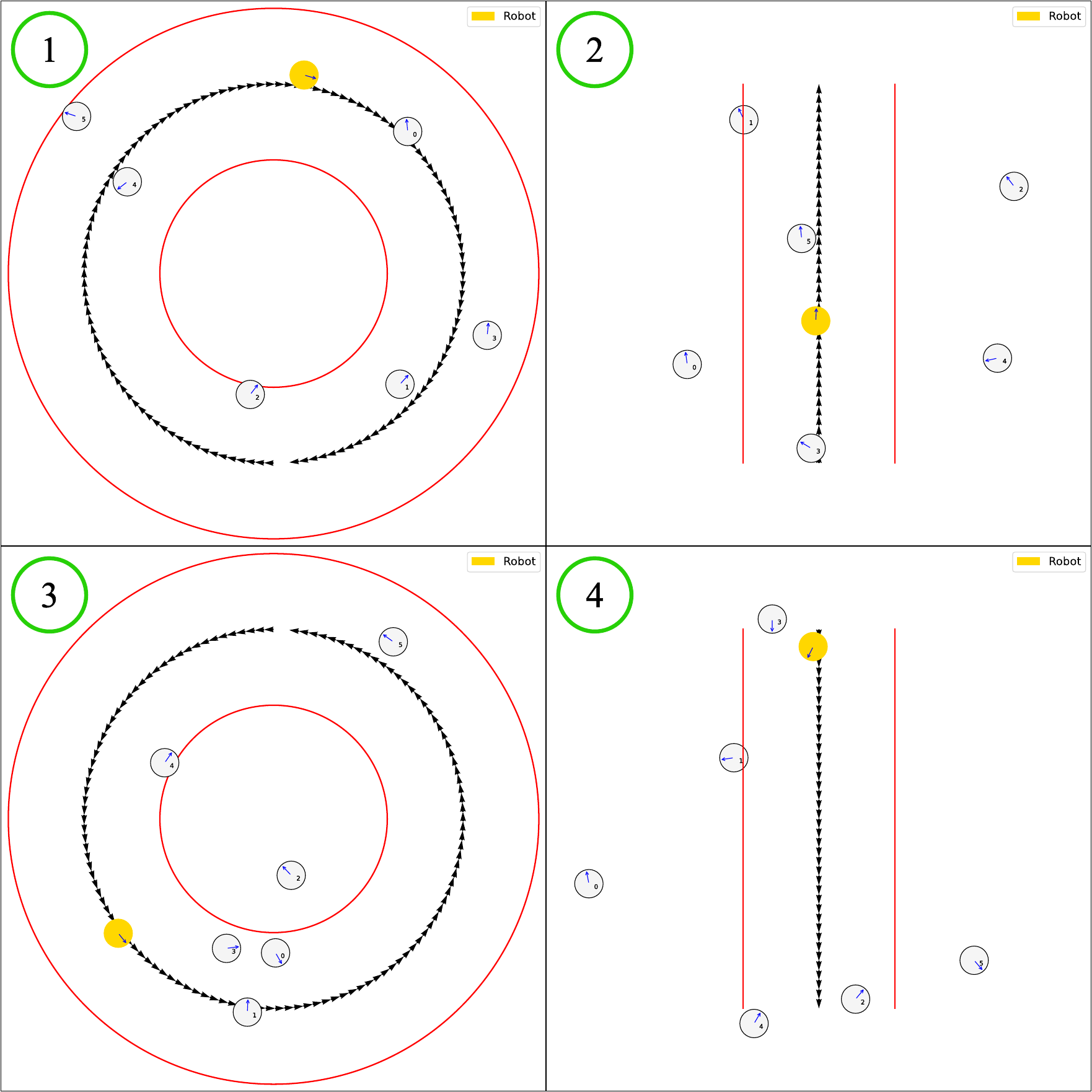}
    \caption{Simulation scenarios. The path is black, the corridors are red, the robot is yellow, and the humans are grey.}
    \label{fig:Training Scenarios}
\end{figure}

Upon detecting an OOD state, we assess the safety of DR-MPC's action using a constant velocity motion model (CVMM). We roll out the robot's proposed action and the human's estimated motion (position difference in the last two timesteps) for two seconds. If a safety-human raise is triggered in the rollout, the action is deemed unsafe. For the safety-corridor condition, we evaluate the one-timestep rollout result.

If the proposed DRL action is both OOD and fails the CVMM safety check, we evaluate a predefined set of alternative actions for safety. We select the safe action that is closest to the MPC action, thereby guiding our DRL agent into regions that are both safe and conducive to path advancement.

\subsubsection{Soft Resets}

In real-world applications, episode resets are inefficient. Thus, unlike traditional episodic RL, we avoid `hard resets’ and perform `soft resets' during training by returning the robot to the closest non-terminal state when a termination condition occurs. This approach of starting at any non-terminal state is analogous to the Monte Carlo Exploring Starts algorithm \cite{SuttonRL}.

\section{Simulation Experiments} \label{sec:Simulation Experiments}
 
\subsection{Simulator and Setup}

We augment the CrowdNav simulator from \cite{IntentionAwareGraph} by replacing waypoint navigation with path tracking. Our training schema cycles through four scenarios (Figure \ref{fig:Training Scenarios}). We run our simulator with $n_t=6$; these 6 humans are modeled using ORCA and continuously move to random goals in the arena. As in \cite{NAX}, because we use a disturbance penalty, the robot is visible to the humans. Our robot has an action space of $v \in [0,1]$ and $\omega \in [-1,1]$. The MDP has a timestep of 0.25s, and we train our models with a limited amount of data: 37,500 steps---around 2.5 hours of data. We set $K=100$ so that by the end of training, over $95\%$ of the states are ID. 

\subsection{Model Baselines}

We evaluate five models. The first applies \cite{IntentionAwareGraph}, which shares a similar backbone to DR-MPC. This model combines the `PT Embedding' and `HA Embedding' with an MLP to generate the agent's action; we refer to this model as the na\"ive DRL model. The second model is residual DRL \cite{ResidualRLForRobotics}, which outputs a corrective action: the action executed in the environment is the sum of the MPC and DRL action truncated into the feasible action space. The corrective action has range $v \in [-1, 1]$ and $\omega \in [-2, 2]$ to ensure the summed action can cover the entire action space. The residual DRL model architecture is the same as the na\"ive DRL model but also has $\mathbf{a}_\text{MPC}$ inputted into the final MLP. We evaluate two versions of DR-MPC: one without the OOD state detection and CVMM modules, and one with them. Finally, we compare against ORCA, which, although it does not optimize for the same objectives as our DRL-based models, serves as an intuitive performance baseline.

\subsection{Results and Discussion}

\begin{figure}
  \centering
  \includegraphics[width=0.5\textwidth]{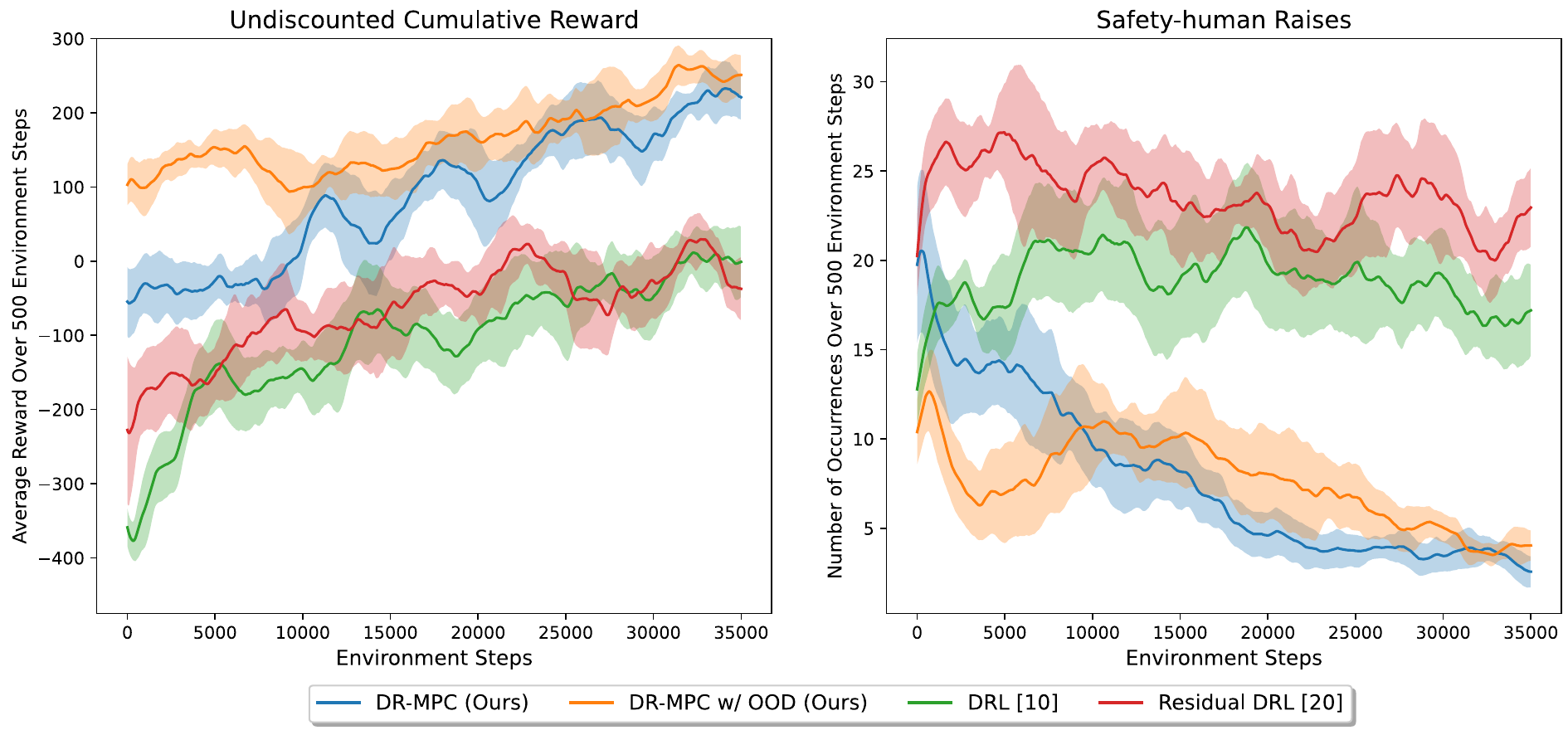}
  \caption{Simulation results averaged over 10 trials. Both DR-MPC models outperform na\"ive DRL and Residual DRL by efficient task switching for human avoidance.
 }
  \label{fig:Sim Results}
\end{figure}

Figure \ref{fig:Sim Results} depicts training progress and Table \ref{tab:performance_metrics} is the final model comparison. From the cumulative reward plot, both DR-MPC models significantly outperform the na\"ive DRL and Residual DRL models, which have difficulty advancing on the path while avoiding safety-human raises. DR-MPC, however, excels in task switching and optimizes both path tracking and human avoidance using the $\boldsymbol{\alpha}$ parameter. ORCA guarantees zero collisions with humans modeled as ORCA and achieves the highest success rate but with the highest NNT. Furthermore, ORCA does not attempt to minimize human comfort (disturbance) or deviation from the path, which is practically useful in the real world where staying closer to the desired path typically aligns better with drivable areas.

As designed, both DR-MPC models start with a high base reward due to their initialization with near-MPC path tracking behaviour, which is better than the residual DRL model, which follows the MPC action only in expectation, resulting in slower convergence and lower initial rewards.

\begin{table}[htbp]
\centering
\caption{Performance Comparison. `SR' is success rate, `EDR' is end deviation rate, `SHRR' is safety-human raise rate, `SCRR' is safety-corridor raise rate, `ATR' is actuation termination rate, and `NNT' is normalized navigation time which is navigation time divided by path length.}
\label{tab:performance_metrics}
\begin{tabular}{lcccccc}
\toprule
\textbf{Model} & \textbf{SR}$\uparrow$ & \textbf{EDR}$\downarrow$ & \textbf{SHRR}$\downarrow$ & \textbf{SCRR}$\downarrow$ & \textbf{ATR}$\downarrow$ & \textbf{NNT}$\downarrow$ \\ 
\midrule
\textbf{ORCA}            & 0.60 & 0.07 & 0.00* & 0.33 & 0.00 & 1.34 \\[4pt] 
\textbf{DRL [10]}        & 0.20 & 0.02 & 0.77 & \textbf{0.01} & \textbf{0.00} & 1.16 \\[4pt] 
\textbf{Residual}        & \multirow{2}{*}{0.21} & \multirow{2}{*}{\textbf{0.00}} & \multirow{2}{*}{0.79} & \multirow{2}{*}{\textbf{0.00}} & \multirow{2}{*}{\textbf{0.00}} & \multirow{2}{*}{\textbf{1.06}} \\ 
\textbf{DRL [20]}        &      &       &       &       &       &       \\[4pt]  
\textbf{DR-MPC}          & \multirow{2}{*}{\textbf{0.57}} & \multirow{2}{*}{0.02} & \multirow{2}{*}{\textbf{0.30}} & \multirow{2}{*}{\textbf{0.00}} & \multirow{2}{*}{0.11} & \multirow{2}{*}{1.32} \\
\textbf{(Ours)}          &      &       &       &       &       &       \\[4pt] 
\textbf{DR-MPC}         & \multirow{3}{*}{\textbf{0.58}} & \multirow{3}{*}{0.06} & \multirow{3}{*}{\textbf{0.31}} & \multirow{3}{*}{\textbf{0.01}} & \multirow{3}{*}{0.04} & \multirow{3}{*}{1.17} \\ 
\textbf{w/ OOD}          &      &       &       &       &       &       \\ 
\textbf{(Ours)}          &      &       &       &       &       &       \\ 
\bottomrule
\end{tabular}
\end{table}

Comparing our DR-MPC models, the one with OOD state detection demonstrates better performance in the early stages of training, as the heuristic policy helps guide it away from collisions. As training progresses and more states become ID, we observe a temporary increase in safety-human raises and a dip in cumulative reward. Shortly after, its performance converges with the model without OOD detection. Thus, OOD detection enables greater initial performance while ultimately achieving similar long-term results.

We qualitatively analyze OOD detection by deploying our trained models in an environment with two additional static humans placed directly on the path. When the robot approaches a static human, OOD detection is triggered, and the heuristic policy guides the robot around the human. In the real-world results (Figure \ref{fig:Real-world Results}), the upward trend of the ID (green) line has a spike at 5000 steps, which is not an artifact but reflects early training challenges in latent representation changes making rare state identification difficult. To overcome this, we start with a stricter KNN distance threshold (we scale the threshold calculation by $\frac{1}{3}$) and gradually relax this scaling factor to identity by the end of training. This tuning leads to better results than a constant distance threshold.

Performance could be further improved with a better heuristic safety check and heuristic policy. Currently, the CVMM safety check is overly conservative: early in training, 21\% of CVMM triggers result in DR-MPC collisions on the next timestep, dropping to 5\% by the end of training. Similarly, early in training when the heuristic policy (which relies on CVMM) causes a collision, DR-MPC also collides 90\% of the time, but this decreases to 3\% by the end of training. These findings suggest that CVMM struggles to model close-contact ORCA behavior. However, the CVMM safety check remains valuable in real-world scenarios, particularly for inattentive humans walking straight or standing still.

\subsection{Reward Ablation}
Given our new decision process formulation, we analyze our reward function. Table \ref{tab:ablation_study} compares the base `DR-MPC w/ OOD' model with all rewards to versions of the model where specific rewards are removed. Interestingly, without $r_\text{pa}$, DR-MPC still achieves a high SR, as the policy begins with MPC path-tracking behavior, allowing it to reach the end of the path; however, this results in a significantly higher NNT. Removing $r_\text{dev}$ increases EDR and SCRR as expected, while removing the goal penalty marginally increases EDR. Without $r_\text{cor-col}^*$, SHRR increases as expected. Removing $r_\text{act}^*$ results in similar overall performance but with higher ATR. Removing $r_\text{hum-col}^*$ causes SHRR to increase drastically. Lastly, removing $r_\text{dist}$ reduces SHRR and greatly increases episodes reaching the path's end (SR + EDR). This result makes intuitive sense because $r_\text{dist}$ influences comfort rather than directly impacting the quantitative navigation metrics. Consequently, while its numerical benefits are evident, the qualitative impact it provides must also be carefully considered. Therefore, we conclude that all rewards are essential to our decision process formulation for real-world social navigation.

\begin{table}[htbp]
\centering
\caption{Reward Ablation Study on DR-MPC w/ OOD}
\label{tab:ablation_study}
\begin{tabular}{lcccccc}
\toprule
\textbf{Model} & \textbf{SR}$\uparrow$ & \textbf{EDR}$\downarrow$ & \textbf{SHRR}$\downarrow$ & \textbf{SCRR}$\downarrow$ & \textbf{ATR}$\downarrow$ & \textbf{NNT}$\downarrow$ \\
\midrule
Base                 & 0.58             & 0.06         & 0.31          & 0.01          & 0.04         & 1.17         \\ 
$r_\text{pa}$            & 0.46             & 0.02         & 0.42          & 0.00          & 0.09         & 1.60         \\ 
$r_\text{dev}$           & 0.46             & 0.08         & 0.19          & 0.06          & 0.21         & 1.27         \\ 
$r_\text{goal}^*$        & 0.56             & 0.07         & 0.37          & 0.01          & 0.00         & 1.17         \\ 
$r_\text{cor-col}^*$     & 0.56             & 0.06         & 0.27          & 0.05          & 0.05         & 1.24         \\ 
$r_\text{act}^*$         & 0.56             & 0.09         & 0.24          & 0.04          & 0.08         & 1.30         \\ 
$r_\text{hum-col}^*$     & 0.24             & 0.00         & 0.76          & 0.00          & 0.00         & 1.07         \\ 
$r_\text{dist}$          & 0.59             & 0.13         & 0.21          & 0.00          & 0.07         & 1.54         \\ 
\bottomrule
\end{tabular}
\end{table}

\vspace{-0.5cm}
\section{Hardware Experiments} \label{sec:Real-world Experiments}

\subsection{Implementation}
We use a Clearpath Robotics Jackal equipped with an Ouster OS0-128 LiDAR that captures $360^{\circ}$ range and reflectivity data. This data can be represented as equirectangular images---an example of the reflectivity image is shown in Figure \ref{fig:Full Pipeline}. These images are comparable to low-resolution cameras, enabling the application of computer-vision models. By fusing the 2D computer-vision results with the LiDAR’s depth information, we can produce 3D outputs. 

For path tracking, we use Teach and Repeat (T\&R) for rapid deployment in new environments \cite{vtr}. Specifically, in LiDAR T\&R, after manually driving the robot through a new environment once, the robot can subsequently localize itself to this previously driven path and track it using MPC.

\begin{figure*}
  \centering
  \includegraphics[width=1.0\textwidth]{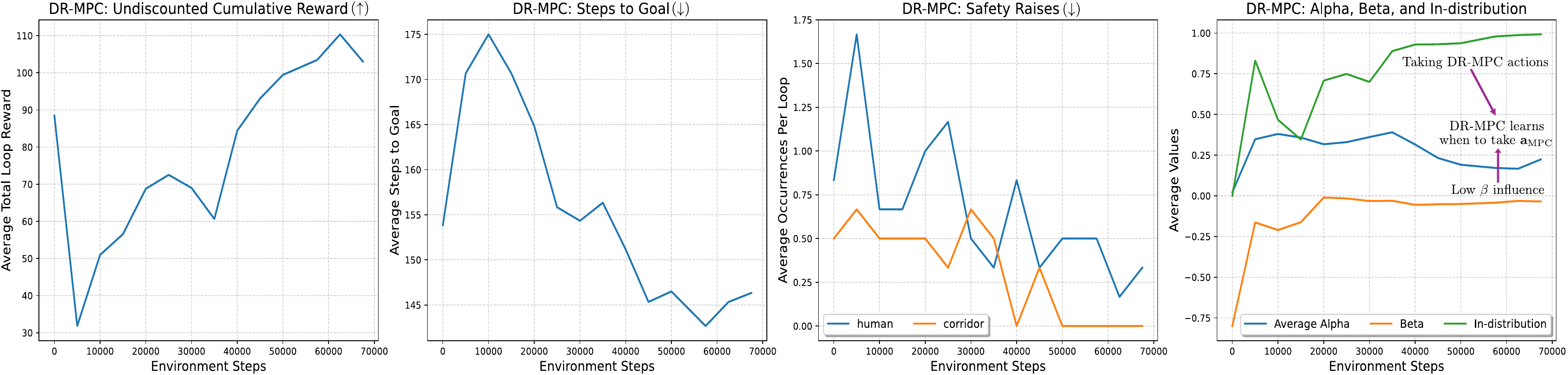}
  \caption{DR-MPC's real-world training results averaged over 6 trials---3 per testing scenario. DR-MPC starts with performance equal to the heuristic policy. However, because of DRL's ability to learn through noise, after exploration, the final model performs better on each key metric compared to where it started.}
  \label{fig:Real-world Results}
\end{figure*}

\begin{figure}
    \centering
    \includegraphics[width=0.49\textwidth]{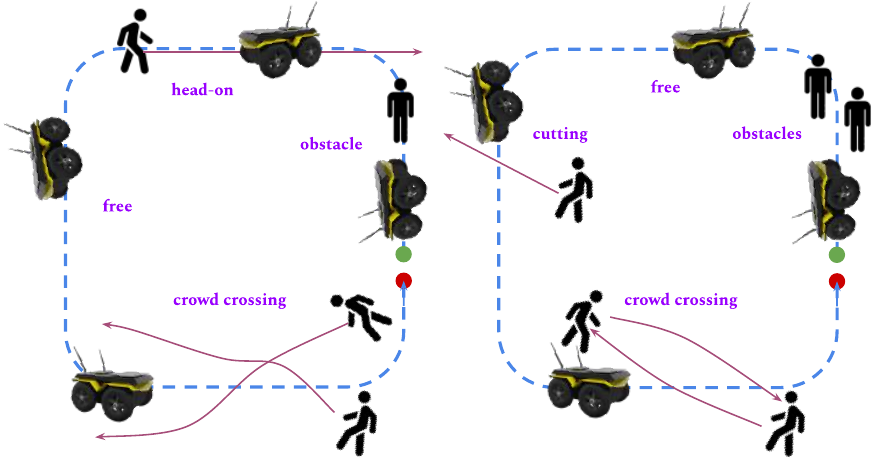}
    \caption{Real-world testing scenarios used to evaluate models. These loops contain diverse situations for social navigation.}
    \label{fig:Testing Scenarios}
\end{figure}

\begin{figure}
  \centering
  \includegraphics[width=0.5\textwidth]{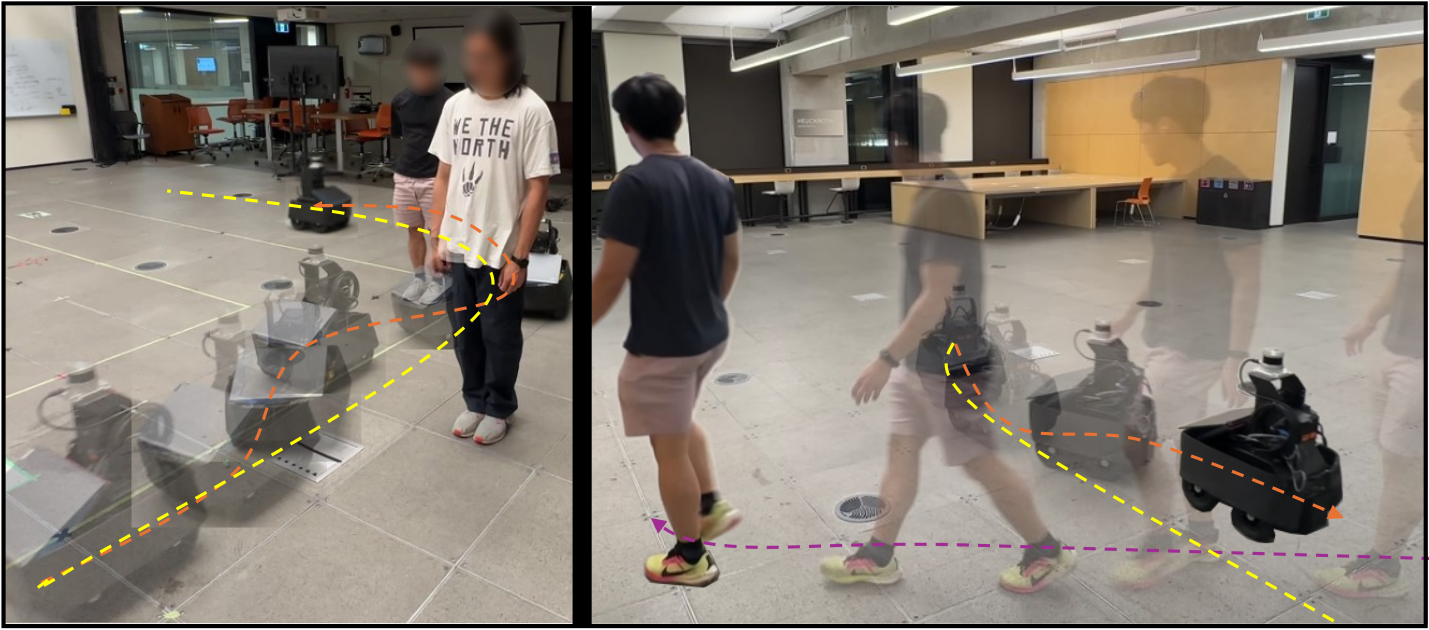}
  \caption{Real-world examples. Reference path is yellow, robot's trajectory is orange, and human's trajectory is purple. Left: robot navigates around two static humans. Right: robot deviates from its path to avoid disturbing the human.}
  \label{fig:RealWorld}
\end{figure}

Lastly, we detect humans by running a pre-trained YOLOX model on the reflectivity image and its $180^{\circ}$ shifted version to account for the wrap-around point \cite{mmdetection}. By combining depth information with the localization result from T\&R, we recover the 3D world positions of humans and track them using the Norfair package \cite{norfair}.

\vspace{-0.5cm}
\subsection{Experimental Setup}
During training, between zero and four humans interact with the robot. The MDP timestep is 0.2s, aligned with incoming LiDAR data every 0.1s. Inference is performed on a ThinkPad P16 Gen 1 with an Intel i7-12800HX Processor and NVIDIA RTX A4500 GPU, while real-time training occurs on another computer. Inference model weights are periodically updated. 

Although we trained on 3.75 hours of experience transitions, the entire process took about 15 hours. After every 500 experience tuples, we paused data collection to allow the model to update, performing twice as many training updates as environment samples. Additionally, outliers caused by processing delays from resource allocation variance were filtered to reduce data noise. Also, soft resets, along with charging the laptop and robot batteries, contributed to the additional time.

We periodically evaluate the model on fixed but diverse scenarios (Figure \ref{fig:Testing Scenarios}). The scenario on the left involves a stationary human, a human walking toward the robot, free driving, and a crowd crossing. The loop on the right tests the robot’s weaving ability between two stationary humans, followed by free driving, a human crossing the robot’s path, and a differently configured crowd crossing. Each model undergoes 6 runs---3 per scenario.

We benchmark real-world DR-MPC against three models: (1) DR-MPC trained in simulation without modification (`Sim Model Raw'), (2) DR-MPC trained in simulation with real-world adjustments, such as observation delays, acceleration constraints, and distance-based detection limits (`Sim Model Adjusted'), and (3) a heuristic policy, which executes $\mathbf{a}_\text{MPC}$ if it passes the CVMM safety check and switches to the heuristic policy used to guide the DR-MPC model if unsafe.

The real world introduces challenges absent in simulation, such as perception errors (missed and false detections), variable processing delays (localization, human detection, action generation), probabilistic human motion, and robot acceleration dependent on battery charge. These challenges highlight a key advantage of DRL: it maximizes expected cumulative reward, allowing effective learning in noisy environments.

\vspace{-0.5cm}
\subsection{Results and Discussion}
\begin{table}[h!]
\centering
\caption{Real-world Policy Results ($\mu \pm \sigma$ \# Per Loop)}
\label{tab:Real-world Final Results}
\resizebox{0.5\textwidth}{!}{%
\begin{tabular}{@{}lccc@{}}
\toprule

\textbf{Algorithm}         & \textbf{Steps to Goal} $\downarrow$ & \textbf{Safety-human}  & \textbf{Safety-corridor} \\ 
                           &                                       & \textbf{Raises} $\downarrow$              & \textbf{Raises} $\downarrow$                \\ \midrule
{Heuristic}         & 158.0 ± 2.7                           & 1.0 ± 0.6                                   & 0.3 ± 0.5                                      \\
{Sim Model Raw}     & 175.3 ± 3.4                           & 0.5 ± 0.5                                   & 3.2 ± 1.1                                      \\
{Sim Model Adjusted}& 161.2 ± 4.7                           & 0.7 ± 0.5                                   & 1.5 ± 0.8                                      \\
{DR-MPC (Ours)}     & \textbf{146.3 ± 3.1}                  & \textbf{0.3 ± 0.5}                          & \textbf{0.0 ± 0.0}                             \\ \bottomrule
\end{tabular}%
}
\end{table}

Table \ref{tab:Real-world Final Results} shows DR-MPC outperforming benchmark models in all key metrics---fewer safety raises and faster navigation. The training process (Figure \ref{fig:Real-world Results}) begins with performance nearly identical to the heuristic policy (as expected). During exploration, DR-MPC performs worse but eventually coalesces its experience to achieve superior results. We notice $\beta$ (orange) trends toward 0, while the average $\alpha$ (blue) remains low, indicating the model learns when to best select the MPC action. After 4 hours of data, DR-MPC's performance begins to plateau. While more data would likely add marginal improvements, social navigation is a long-tail problem, and even in simulation with deterministic humans, safety-human raises never reach zero.

Qualitatively, DR-MPC exhibits key crowd-navigation behaviors: smoothly weaving around still humans (Figure \ref{fig:RealWorld} left) and deviating from its path and slowing down for moving humans (Figure \ref{fig:RealWorldMoneyShot} and Figure \ref{fig:RealWorld} right). In human-free areas, DR-MPC switches to MPC path tracking.

The heuristic policy performance suffers because human velocity estimates are noisy, as the position is inferred from the bounding box center, which shifts based on body orientation. Also, this policy does not account for delays or the robot's acceleration, reducing its ability to navigate around humans. However, it still guides DR-MPC towards open spaces.

`Sim Model Raw' performs poorly, oversteering and colliding with virtual boundaries due to unaccounted state delays. Adjusting the simulator (`Sim Model Adjusted') improves performance but still falls short of DR-MPC trained in the real world. This underscores the advantage of real-world training in handling noisy data and developing robust policies.

\vspace{-0.5cm}
\subsection{Limitations and Future Work}
One limitation is the manual tuning of the heuristically defined reward function, particularly balancing path-tracking and human-avoidance rewards. For instance, overemphasizing collision penalties impedes progress along the path; Imitation Learning methods bypass this limitation by replicating expert behaviour \cite{SCAND}. Future work includes incorporating reward learning methods, such as Inverse RL \cite{IRLForSocialNavigation} and preference learning \cite{NaviSTAR}, to enable DR-MPC to acquire more human-aligned behaviors.

Another limitation involves false collision detections due to modeling humans as circles. False positives occur because humans are often wider shoulder-to-shoulder than front-to-back. False negatives arise when extended feet during walking or resting postures cause the center point to misrepresent their state. Future work includes skeleton detection to better capture nuances, improving collision accuracy.

Lastly, we acknowledge that our `real-world' experiments capture only a subset of human behaviors expected in true `in-the-wild' scenarios. Our results reflect interactions with people familiar with the robot. Behaviors such as stopping to observe the robot or intentionally obstructing its path are not encompassed in our current environment. In future work, we aim to conduct `in-the-wild' experiments to evaluate the robot's performance with more diverse interactions.

\section{Conclusion}
We introduced DR-MPC, a novel integration of MPC path tracking with DRL, and demonstrated its effectiveness and superiority to prior work in both simulation and real-world scenarios. Training a DRL agent directly in the real world bypasses the sim-to-real gap, addressing the inevitable mismatch between the dynamics of modeled humans and real humans. While simulation is crucial for model development, the ultimate goal is deploying DRL agents that perform effectively in real-world conditions, where a plethora of challenges remain.

\section*{Acknowledgment}
James would like to thank Alexander Krawciw for his invaluable mentorship in working with real robots. James is supported by the Vector Scholarship in AI and the Queen Elizabeth II Graduate Scholarship.

\bibliographystyle{IEEEtran}
\bibliography{ref}

\appendices

\section{Human Avoidance Network}

Compared to \cite{IntentionAwareGraph}, we use the humans' past trajectories rather than their forecasted trajectories. This way, DRL directly learns from the sensor noise embedded in the state. 

For each human trajectory $\mathbf{q}^{t-H_{i}:t}_{i} = \begin{bmatrix}
    \mathbf{p}^{t-H_i}_i \dots \mathbf{p}^t_i
\end{bmatrix}$, we sequentially process it from time $t-H_{i}$ to $t$ using a GRU to generate an embedded trajectory $\mathbf{e}^i_\text{traj}$. This step handles human trajectories of varying lengths, embedding them into a uniform latent space. Next, with $\mathbf{E}_\text{traj} = \begin{bmatrix}
    \mathbf{e}^1_\text{traj} \dots \mathbf{e}^{n_t}_\text{traj}
\end{bmatrix}$, we use three MLPs to generate the queries $\mathbf{Q}_\text{traj}$, keys $\mathbf{K}_\text{traj}$, and values $\mathbf{V}_\text{traj}$. We then apply multi-head attention using the scaled dot-product attention: $\text{MultiHead}(\mathbf{Q}_\text{traj},\mathbf{K}_\text{traj}, \mathbf{V}_\text{traj})$. The output of this module is $\mathbf{E}_\text{HH}$, a $n_t \times d_{HH}$ tensor, where $d_{HH}$ is the dimension of the human-human embeddings.

Next, we process the robot trajectory $\mathbf{r}^{t-H:t}$ by embedding it with an MLP into $\mathbf{e}^{\text{robot}}_\text{traj}$. We then compute the robot-human attention. Here, the keys $\mathbf{K}_{robot}$ are generated from $\mathbf{e}^{\text{robot}}_\text{traj}$ and the queries $\mathbf{Q}_{\text{HH}}$ and the values $\mathbf{V}_\text{HH}$ from $\mathbf{E}_\text{HH}$. The result of this multi-head attention network is the embedding $\mathbf{e}_{RH}$.

Finally, we concatenate $\mathbf{e}_{RH}$ with $\mathbf{v}^{t-H:t-1}$ and pass this tensor through one last MLP to obtain the crowd embedding $\mathbf{e}_{HA}$, which is then used to generate the 6 mean actions for human avoidance.

\end{document}